\documentclass[letterpaper, 10 pt, conference]{ieeeconf}  

\IEEEoverridecommandlockouts                              

\usepackage{hyperref}
\usepackage{graphics} 
\usepackage{amsmath} 
\usepackage{amssymb}  
\usepackage{cite}
\usepackage{algorithm}
\usepackage{booktabs}
\usepackage{algorithmicx}
\usepackage{cleveref}
\usepackage{algpseudocode}
\usepackage{graphicx}
\usepackage{mathtools}
\usepackage{xcolor}
\usepackage{soul}
\usepackage{bbm}
\usepackage{makecell}
\usepackage{subcaption}
\usepackage{cancel}

\newcommand{\ind}{\perp\!\!\!\!\perp} 

\newcommand{\kroneckerDelta}{\delta}

\newcommand{\stateSpace}{\mathcal{S}}

\newcommand{\dynamicsFunction}{\mathcal{D}}
\newcommand{\transitionFunction}{T}
\newcommand{\observationFunction}{O}

\newcommand{\dynamicsPrior}{p_{\dynamicsFunction}}
\newcommand{\transitionPrior}[1]{p(\transitionFunction_{#1})}

\newcommand{\param}{\theta}
\newcommand{\paramSpace}{\Theta}
\newcommand{\paramPrior}{\param_0}
\newcommand{\gbapomdpDynamics}{\bar{\mathcal{D}}}
\newcommand{\parameterUpdate}{\mathcal{U}}

\newcommand{\physicsSimulator}{\transitionFunction_r}
\newcommand{\knownObservationComponent}{\observationFunction_r}
\newcommand{\knownObservationVariable}{o_r}

\newcommand{\humanDynamics}{\transitionFunction_h}
\newcommand{\humanPrior}{\transitionPrior{h}}
\newcommand{\unknownObservationComponent}{\observationFunction_h}

\newtheorem{definition}{Definition}

\title{\LARGE \bf
On-Robot Bayesian Reinforcement Learning for POMDPs
}

\author{Hai Nguyen$^{1*}$, Sammie Katt$^{1}$, Yuchen Xiao$^{1}$, Christopher Amato$^{1}$%
\thanks{$^{1}$Khoury College of Computer Sciences, Northeastern University, Boston, MA 02115, USA. $^{*}$Corresponding author \texttt{nguyen.hai1@northeastern.edu}.}
}

\begin{document}

\maketitle
\thispagestyle{empty}
\pagestyle{empty}

\begin{abstract}
Robot learning is often difficult due to the expense of gathering data.
The need for large amounts of data can, and should, be tackled with effective algorithms and leveraging expert information on robot dynamics.
Bayesian reinforcement learning (BRL), thanks to its sample efficiency and ability to exploit prior knowledge, is uniquely positioned as such a solution method.
Unfortunately, the application of BRL has been limited due to the difficulties of representing expert knowledge as well as solving the subsequent inference problem. This paper advances BRL for robotics by proposing a specialized framework for physical systems. In particular, we capture this knowledge in a factored representation, then demonstrate the posterior factorizes in a similar shape, and ultimately formalize the model in a Bayesian framework. We then introduce a sample-based online solution method, based on Monte-Carlo tree search and particle filtering, specialized to solve the resulting model. This approach can, for example, utilize typical low-level robot simulators and handle uncertainty over unknown dynamics of the environment.
We empirically demonstrate its efficiency by performing on-robot learning in two human-robot interaction tasks with uncertainty about human behavior, achieving near-optimal performance after only a handful of real-world episodes. A video of learned policies is at \url{https://youtu.be/H9xp60ngOes}.
\end{abstract}


\section{INTRODUCTION}

Mainstream reinforcement learning (RL) techniques~\cite{mnih2015human, mnih2016asynchronous, schulman2017proximal} are not sample efficient enough for online applications in physical systems: long training hours or unguided exploration can wear out or break fragile, expensive robot systems. Instead, the common approach for policy learning in robotics is to use simulators, and transfer learned policies to the real hardware (\emph{sim2real}). However, simulators cannot capture the real world exactly; therefore, additional techniques are required for successful transfers, such as online system identification~\cite{yu2017preparing} or domain randomization~\cite{ peng2018sim}. These approaches have shown success, but the research question is far from solved, and transfers fail when the sim2real gap is large. Moreover, none of these approaches can, in principle, exploit prior (expert) knowledge, which is useful for faster learning and often exists in most robotics problems. In our view, the ideal approach should learn directly on physical hardware (\emph{on-robot}) while leveraging prior knowledge.

Bayesian Reinforcement Learning (BRL) has great potential for on-robot learning. 
BRL is sample-efficient and provides a principled solution to the exploration-exploitation trade-off by explicitly incorporating uncertainty into the decision-making process. 
Furthermore, BRL allows expert knowledge to be easily integrated into the learning process as ``priors'', allowing the agent to exploit knowledge it would otherwise have needed many precious samples to learn. 
Despite its promise, previous attempts at BRL for on-robot learning have been limited, even for the fully observable settings. 
This can be explained by the scaling issues of BRL methods and the difficulty of representing the complex expert knowledge available in physical systems.
In particular, these methods typically assume straightforward priors, such as a single large Dirichlet table or neural networks~\cite{katt2022baddr,ross2011bayesian}, but rarely provide tools to tackle more sophisticated applications.

This paper presents a Bayesian approach incorporating expert knowledge for efficient on-robot learning under partial observability. 
We first identify typical and reasonable assumptions in physical systems, including some inspired by mixed observability Markov decision processes~\cite{ong2009pomdps}, and translate them to a Bayesian setting.
The next step derives the corresponding Bayesian inference problem, showing how the nature of the prior that we discovered shapes the posterior over the unknown quantities of the problem.
This leads to the formalization of a Bayes-adaptive (BA) model specialized for robotics systems.
Finally, we propose a method for solving this model by specializing Monte-Carlo tree search and particle filtering, which provides an efficient and principled solution to the original learning problem.
We empirically demonstrate its efficiency with on-robot learning in two human-robot interaction tasks with uncertainty about human behavior, achieving near-optimal performance after only a handful of real-world episodes.

\section{RELATED WORK}

\subsection{Bayesian RL under Partial Observability}

Bayesian RL involves maintaining a probability distribution over an augmented state, including both the environment dynamics and the environment state, as proposed by the BA-POMDP framework~\cite{ross2011bayesian}. State-of-the-art BRL methods~\cite{katt2017learning, katt2019bayesian, katt2022baddr} for partially observable domains have improved on the BA-POMDP by incorporating the Partially Observable Monte Carlo Planning (POMCP)~\cite{silver2010monte}, an efficient online planner for large POMDPs. While earlier methods~\cite{katt2017learning, katt2019bayesian} are only applicable to small discrete POMDPs or ones with factorizable dynamics, Bayes-adaptive deep dropout RL (BADDr)~\cite{katt2022baddr} was introduced to tackle more complex domains. BADDr leverages the representation power of dropout networks~\cite{srivastava2014dropout} to capture priors, making BRL scalable to larger domains while being considerably more sample-efficient than pure RL methods. In this paper, we also use dropout networks to construct (parts of) a specialized dynamics model, leveraging assumptions about the factored dynamics and full observability of the robot state.

\subsection{On-Robot Reinforcement Learning}
Recently, many methods have performed learning directly on physical hardware. In~\cite{gu2017deep}, multiple robot workers are used to collect data for online training of a robot arm to open a door autonomously.~\cite{singh2019end} used human feedback to learn manipulation tasks in 1-4 hours. Meanwhile,~\cite{zhan2020framework} combined RL and contrastive learning~\cite{oord2018representation} to learn manipulation tasks online from pixels. In addition, \cite{wang2022robot, zhu2022sample} used symmetry-aware neural networks to encode domain symmetries to perform online learning manipulation tasks within several hours, while~\cite{smith2022walk, wu2022daydreamer} conducted online learning for locomotion tasks on legged robots. Nevertheless, unlike ours, none of these works addresses partially observable domains or employs a Bayesian approach directly to a physical system.

\section{BACKGROUND}

\subsection{Partially Observable MDP (POMDP)}
A POMDP~\cite{kaelbling1998planning} is formally defined by the tuple $(\mathcal{S}, \mathcal{A}, \Omega, T, O, R, p_{s_0})$, where $\mathcal{S}, \mathcal{A}, \Omega$ are respectively the state, action, and observation spaces; $T: \mathcal{S} \times \mathcal{A} \rightarrow \Delta(\mathcal{S}), O: \mathcal{S} \times \mathcal{A} \rightarrow \Delta(\Omega)$, and $R: \mathcal{S} \times \mathcal{A} \rightarrow \mathbb{R}$ are respectively the transition, observation, and reward functions; $p_{s_0} = \Delta(\mathcal{S})$ is the prior about the initial state $s_0$. The goal is to find a sequence of actions to maximize the discounted return defined as $\mathbb{E}[\sum_{t=0}^\infty \gamma^t r_t]$~\cite{sutton2018reinforcement} with some discount factor $\gamma \in [0, 1]$. An action $a$ taken in state $s$ will result in a state $s'$, sampled from the transition function $T(s, a, s') = p(s'\mid s, a)$. In a POMDP, the agent can only observe $o \in \Omega$, indirectly related to $s'$ via the observation function $O(s', a, o) = p(o \mid s', a)$. This implies that the agent might need to use the entire action-observation history, which grows with the episode length, to act optimally. Alternatively, acting optimally can rely on a belief (posterior probability distribution) $b \in \Delta(\mathcal{S})$ over possible states, where $b(s)$ denotes the probability that the environment's true state is $s$. Given a new action $a$ and observation $o$, a new belief $b'$ can be calculated as:
\begin{align}
    b'(s') \propto O(s', a, o) \sum_{s \in \mathcal{S}} T(s, a, s') b(s).
    \label{eq:belief_update}
\end{align}
\Cref{eq:belief_update} indicates that explicitly belief tracking (and therefore planning) must rely on access to the \emph{known} dynamics $T(s, a, s')$ of the system. When these are unknown, we must turn to learning-based approaches instead. This work adopts the \emph{Bayesian} perspective.

\subsection{General Bayes-Adaptive POMDP (GBA-POMDP)}\label{sec:gba-pomdp}

BRL assumes that, instead of access to the system's dynamics $\mathcal D$, we have a parameterized \emph{prior} probability distribution $p(\mathcal D; \param)$.
On a high level, our agent will maintain a belief $\bar b \in \Delta(S, \paramSpace)$ over the dynamics and the state and pick actions that optimize future return \emph{with respect to this belief}. 
Technically, this Bayesian framework constructs a ``belief'' POMDP: an augmented POMDP of which the state space (and dynamics) include parameters that describe the dynamics of the underlying system. 
Even if the dynamics of the original system are unknown, it is now again possible to plan and track beliefs in the larger POMDP instead. 

\begin{definition}[GBA-POMDP]
Given a dynamics prior $\param \in \paramSpace$, and a parameter update function $\mathcal{U}$, then a general BA-POMDP is a POMDP defined by the tuple $(\bar{\mathcal{S}}, \mathcal{A}, \Omega, \gbapomdpDynamics, O, R, p_{{\bar s}_0})$ with augmented state space $\bar{\mathcal{S}} = S \times \paramSpace$ and prior $p_{{\bar s}_0} = (p_{s_0}, \paramPrior)$.
Denote $\kroneckerDelta_x$ as the Kronecker-delta function, which is zero everywhere except at $x$, then the update function $\mathcal{U}$ determines the augmented dynamics model $\gbapomdpDynamics(s', \param', o \mid s, \param, a)$, specified as:
\begin{equation}
    \gbapomdpDynamics = p(s', o \mid s, a; \param) \kroneckerDelta_{\param'}(\parameterUpdate(\param, s, a, s', o))\,,
\end{equation}
where $ p(s', o \mid s, a; \param)$ is the dynamics according to a model parameterized by $\param$.
\end{definition}

Since the GBA-POMDP \emph{is a} POMDP, it can be solved through belief tracking and planning.
Unfortunately, the state space is very large, and we need to reach for approximation techniques instead, which will be covered next.

\subsubsection{Belief Tracking}

\begin{algorithm}
\caption{Online Belief Tracking $(b, a, o, P)$}\label{alg:belief-tracking}
\begin{algorithmic}[1]
\Require belief $b = \{ s, \param \}^P$, action $a$, and observation $o$
\State $b' \leftarrow \emptyset$ \Comment{\textcolor{blue}{Empty next belief}}
\While {sizeof$(b') < P$}
\State $(s, \param) \sim b$ \Comment{\textcolor{blue}{Sample augmented state}}
\State $(s', \param', \tilde{o}) \sim \gbapomdpDynamics(s, \param, a)$ \Comment{\textcolor{blue}{Use GBA-POMDP dynamics}}
\If{$\tilde{o} = o$} \Comment{\textcolor{blue}{Compare with real observation}}
\State {Add $(s', \param')$ to $b'$}
\EndIf
\EndWhile\\
\Return $b'$
\end{algorithmic}
\end{algorithm}

Belief tracking is approximated with particle filtering, in this case, rejection sampling (\cref{alg:belief-tracking}).
Given a current belief $b = p(s, \param)$ and new action-observation pair $(a,o)$, rejection sampling repeatedly samples, updates, and accepts/rejects particles.
In particular, each iteration samples an augmented state from the belief $(s, \param) \sim b$ and proposes a next augmented state using the dynamics $(s', \param', \tilde{o}) \sim \gbapomdpDynamics(s, \param, a)$.
The proposal is accepted and added to the new belief when the simulated observation matches the \emph{real} observation $\tilde{o} = o$, and otherwise rejected.
The process repeats until we have $P$ particles to represent $b'$.
Note that the updated parameters $\param$ are \emph{preserved} throughout (not reset in-between episodes) so that the dynamics are continuously learned through episodes.

\subsubsection{Online Planning}

\Cref{alg:pomcp} shows how Partially Observable Monte Carlo Planning (POMCP)~\cite{silver2010monte}, a Monte-Carlo tree search method, is used for action selection by building a look-ahead tree to evaluate the expected return of each action. The tree is incrementally built through simulations by calling a \emph{Simulate} function $N_s$ times, each starting with a sampled augmented state from the current belief (represented by $P$ particles). When the tree is completed, the action with the largest value is selected. At the next timestep of action selection, the tree will be discarded and built anew. For more details on modifying POMCP for GBA-POMDP, please refer to~\cite{katt2017learning}.

\begin{algorithm}
\caption{POMCP $(b, N_s, P)$}\label{alg:pomcp}
\begin{algorithmic}[1]
\Require Current belief $b = \{ s, \param \}^P$
\Require Number of simulations $N_s$
\State $h_0 \leftarrow \emptyset$ \Comment{\textcolor{blue}{Empty history}}
\For {$i=1 \rightarrow N_s$}
\State {$(s, \param) \sim b$} \Comment{\textcolor{blue}{Sample augmented state}}
\State Simulate$(s, \param, h_0)$ \Comment{\textcolor{blue}{Build a tree}}
\EndFor
\State Return $\operatorname*{argmax}_{b}Q(h_0b)$ \Comment{\textcolor{blue}{Greedy action selection}}
\end{algorithmic}
\end{algorithm}

\subsection{Bayes-Adaptive Deep Dropout RL (BADDr)}\label{sec:baddr}

The previous section described the GBA-POMDP as a recipe that constructs a belief POMDP from a prior $\paramPrior$ and parameter update $\parameterUpdate$.
BADDr~\cite{katt2022baddr} is a realization of GBA-POMDP that uses dropout neural networks~\cite{srivastava2014dropout} to represent the dynamics.
In particular, it assumes that the prior over the dynamics is captured by a parameter set $w \in \mathcal W$ (\emph{e.g.}, $w$ can be neural network weights) and chooses the update function to be stochastic gradient descent with dropout ($\text{SGD}$)~\cite{robbins1951stochastic}, which can be interpreted as an approximation of Bayesian inference over neural network parameters~\cite{gal_dropout_2016}. Consequently, 
BADDr is a POMDP with augmented state space $\bar\stateSpace = \stateSpace \times \mathcal{W}$ and dynamics:
\begin{align}
    \bar\dynamicsFunction(\bar s', o \mid \bar s, a) &= p(s', o \mid s, a; w) \kroneckerDelta_{w'}(\parameterUpdate(w, s, a, s', o))\,, \label{eq:gbapomdp-dynamics} \\
    \parameterUpdate &= w - \nabla \mathcal{L}((s, a), (s',o);w)\,,\label{eq:update}
\end{align}
where $\mathcal{L}((s, a), (s',o);w) = -\log p(s', o \mid s, a; w)$ is the cross-entropy loss between the predicted and true next state and observation.

\subsubsection{Prior Construction}

The initial state distribution of BADDr $p_{\bar s} \in \Delta(S, \dynamicsFunction)$ is the product of the (given) initial POMDP state distribution $p_s$ and a prior over the dynamics.
BADDr parameterizes this prior $\dynamicsPrior$ with parameters $w_T$ and $w_O$, representing the transition and observation models, respectively. 
In general, it is unclear how to set network parameters to reflect prior knowledge, so a data-centric approach is taken instead.
In particular, we sample POMDPs from the (assumed given) prior over the dynamics $p_\dynamicsFunction$, which in turn are used to generate transitions $(s, a, s', o)$ with, for example, a random policy. 
The prior networks $w_T$ and $w_O$ are trained using~\cref{eq:update} on transitions generated from the simulators. 
After training, $\dynamicsPrior$ is represented with an ensemble of $N$ parameter sets $\{(w_T, w_O)\}^N$.

\subsubsection{Online Adaptation}
A solution to BADDr is, as in any (GBA-)POMDP, a combination of belief tracking and online planning.
After picking an action with POMCP and receiving observation, the agent updates its belief over both the current POMDP state and its dynamics according to BADDr's dynamics.
As a result, the belief over (the dynamics) parameters $w_T$ and $w_O$ are updated like in~\cref{eq:update}.
This time, however, uses a \emph{single} sample with the \emph{real} observation from the environment (line 4-6 in~\cref{alg:belief-tracking}).

\section{Baysian Inference in Robotic Systems}

The Bayes-adaptive models described in the background are powerful in their ability to capture a broad class of problems. 
In practice, however, we have \emph{more intricate prior knowledge than previous literature can capture}.
For example, we may be confident enough in our understanding of the sensors that there is no need to learn a model of them or have a reliable low-level (physics) simulator.
On the other hand, certain aspects of the real world will be unknown, such as preferences of collaborating humans, and encoding prior knowledge (that can be updated during execution) over them will be important. 
This knowledge can not be expressed with, for example, BADDr, which assumes a single set of parameters to describe the prior as a distribution.
Similarly, most robots come with a reliable low-level (physics) simulator, which would be unrealistic to try to represent with Dirichlet distributions or even neural networks. 
In short, there is a need to incorporate mixed and partial prior knowledge from real-world applications to Bayesian methods.

\subsection{Prior Knowledge in Robotic Systems}\label{sec:robotic-priors}

Typical assumptions include:

\begin{itemize}
    \item There is a high-fidelity simulator of the state of the robot
    \item The (internal) state of the robot is fully observable
    \item Some high-level system behavior is unknown
    \item The model of the sensors is somewhat known
\end{itemize}

Expert knowledge in robotic tasks typically includes a physical simulator of the robot in which we can simulate actions.
Similarly, the internal of the robot state is observable in most systems. 
What is likely unknown, and over which we may only have a distribution, is some high-level element of the dynamics of the environment in which the robot will be deployed. 
In this work, this is the human behavior in a human-robot setting.
Other assumptions may apply for different applications, but we hope that most classes of assumptions have been covered here and that it will be relatively straightforward to adapt to future requirements.

\begin{figure}[htbp]
    \centering
    \includegraphics[width=1.0\linewidth]{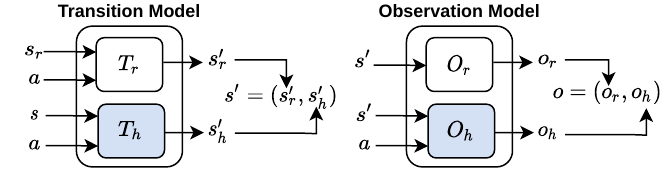}
    \caption{By assuming that part of the dynamics $T_r(s_r, a, s'_r)$ is known and part of the state ($\knownObservationVariable$) computed from the state through $O_r$ is fully observable, we only need to learn smaller transition ($T_h$) and observation ($O_h$) models (blue boxes).}
    \label{fig:reduction}
\end{figure}

We observe that, without loss of generalization, the state can be divided into internal (robot $s_r$) and external (here ``human's'' $s_h$) parts: $s = (s_r, s_h)$ (see~\cref{fig:reduction}).
Then we denote $\physicsSimulator(s_r, a, s'_r)$ as the accurate but low-level simulator in which we can accurately predict the \emph{agent's internal state} by unrolling a controller. 
On the other hand, the dynamics of the environment $\humanDynamics(s, a, s'_h)$ are unknown before deployment.

We make a similar distinction in the observation and divide it into an internal (robot) $\knownObservationVariable$ and external (human) $o_h$ part $o = (\knownObservationVariable, o_h)$. 
We assume that the internal state $\knownObservationVariable$ (can be computed from the state through a given function $O_r$: $s' \rightarrow \knownObservationVariable$) is fully observable (note that a similar assumption is formalized in the \emph{mixed observability} MDP~\cite{ong2009pomdps}).
The observation function over the variable $o_h$, $\unknownObservationComponent(s', a, o_h)$, may or may not be given depending on our understanding of the robot's sensors.

\subsection{Factored Bayesian Inference}\label{sec:factored-priors}

The GBA-POMDP does not provide the tools to capture the expert knowledge we discussed above or how to do further inference over.
Instead, we derive a more sophisticated framework starting from the assumptions stated above.

\paragraph{Priors}

For simplicity, let us consider only the state transition model and note that the state space consists of features $\stateSpace = (\stateSpace_r, \stateSpace_h)$.
According to our assumptions, the transition model $\transitionFunction$ can be factorized into each feature:
$\transitionFunction(s, a, s') = \physicsSimulator(s_r, a, s'_r) \humanDynamics(s, a, s'_h)$.
Indeed, our prior knowledge is factored this way: $\transitionPrior{} = \transitionPrior{r}\transitionPrior{h}$.
Specifically, the dynamics over the internal robot state are assumed known (\emph{i.e.}, a physics simulator $f$), represented with the Kronecker-delta distribution $p(T_r) = \kroneckerDelta_f(T_r)$.

\paragraph{Posteriors}

In the interest of updating our belief over the dynamics, we consider the shape of the posterior $p(\transitionFunction | \{s,a,s',o\})$.
Given the transition $sas'o = \{s, a, s', o\}$, we first apply Bayes' rule and notice an interesting but crucial property:
\begin{align}
    p(\physicsSimulator, \humanDynamics \mid sas'o) &= p(\physicsSimulator \mid sas'o) p(\humanDynamics \mid \cancel{\physicsSimulator}, sas'o) \nonumber \\
    &= p(\physicsSimulator \mid sas'o) p(\humanDynamics \mid sas'o). \label{eq:posterior}
\end{align}
\Cref{eq:posterior} indicates that the transition functions are independent of each other given the transition, $\humanDynamics \ind \physicsSimulator | sas'o$.
This, first, implies that the posterior is also factored.
Second, as a result, Bayesian inference can be made in separate computations, one for each factor of the prior!
In this instance, due to knowing $\physicsSimulator$, the posterior simplifies to $\kroneckerDelta_f(\physicsSimulator)p(\humanDynamics \mid sas'o)$, where the last term depends on the prior representation $\humanPrior$.

\subsection{Formal Definition as a Specialized Bayesian Model}

Here we leverage the shape of the posterior derived above to define a Bayesian framework for robotics. 
In particular, we use the factorization of the posterior computation to construct a model with a factorization of the parameter update rule $\parameterUpdate$ --- one for each factor.

The transition posterior computation, for example, is defined by an update rule for the distribution over the physics simulator and one for the distribution over the human dynamics.
The update rule that corresponds to the Kronecker-delta function is simply its identity $\parameterUpdate(\physicsSimulator, s, a, s', o) = \physicsSimulator$.
The posterior parameters of the human dynamics depend on the choice of the prior model $p(\humanDynamics)$, which in this case will be neural networks $w_{hT}$ with the corresponding dropout SGD update rule $\parameterUpdate(w_{hT}, s, a, s', o)$ as in~\cref{eq:update}.
For example,~\cref{eq:posterior} can be (re-)written with a factored parameter update rule:
\begin{align}
    \cref{eq:posterior} &= \parameterUpdate(\kroneckerDelta_f(\physicsSimulator),s,a,s',o) \parameterUpdate(w_{hT}, s, a, s', o) \nonumber \\
    &= \kroneckerDelta_f(\physicsSimulator)\text{SGD}(w_{hT},s,a,s',o) \label{eq:factored-parameter-update}\,,
\end{align}
and we apply the same approach to the observation function.

The prior of known components (\emph{e.g.}, $\physicsSimulator$ and $O_r$) are, as discussed before, encoded by Kronecker-delta distributions.
Priors over other parts of the dynamics, such as the human's dynamics $\humanDynamics$ and observations $\unknownObservationComponent$, are a design choice, and here are assumed to be represented by some network parameters $(w_{hT}, w_{hO})$.

\begin{definition}
Formally, the resulting Bayes model is a POMDP with state space $\bar\stateSpace = (\stateSpace_r, \stateSpace_h, \mathcal{W}_{hT}, \mathcal{W}_{hO})$ and dynamics:
\begin{align}\label{eq:main}
    \bar\dynamicsFunction(s_r', s_h',& w_{hT}', w_{hO}', o_r, o_h \mid s_r, s_h, w_{hT}, w_{hO}, a) = \\
    (\text{state}) \hspace{1em} & p(s_r' \mid s_r, a; \physicsSimulator) p(s'_h \mid s, a; w_{hT}) \times  \nonumber \\ 
    (\text{obs}) \hspace{1em} & p(\knownObservationVariable \mid s'; \knownObservationComponent) p(o_h \mid s', a; w_{hO}) \times \nonumber \\
    (\text{params}) \hspace{1em} & \kroneckerDelta_{w_{hT}'}(\parameterUpdate_{hT}(w_{hT}, s, a, s'_h)) \times \nonumber \\ 
    & \kroneckerDelta_{w_{hO}'}(\parameterUpdate_{hO}(w_{hO}, s', a, o_h)) \nonumber\,,
\end{align}
where the updates of the known components are omitted for brevity, and those of the neural network parameter are:
\begin{align}
    \parameterUpdate_{hT} &= w_{hT} - \nabla \mathcal{L}((s, a), s_h';w_{hT}) \\
    \parameterUpdate_{hO} &= w_{hO} - \nabla \mathcal{L}((s', a), o_h);w_{hO}).
\end{align}  

\end{definition}

\begin{figure*}[htbp]
    \centering
    \includegraphics[width=0.8\linewidth]{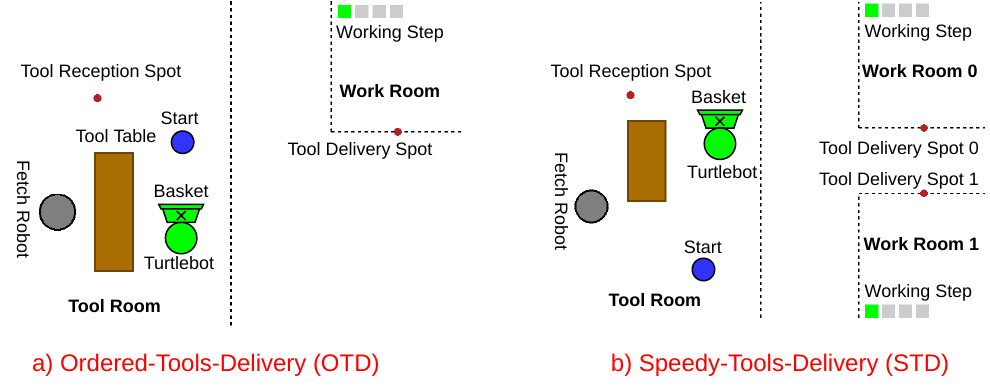}
    \caption{Two human-robot interaction domains are used in our experiments.}
    \label{fig:domain}
\end{figure*}

\subsection{Solving the Bayesian Model}

Starting with the problem definition of a general robotic system, we have developed a belief POMDP that captures typical expert prior knowledge and allows for Bayesian inference over unknown parts of the dynamics.
To solve the resulting problem, however, we require efficient action selection and belief approximation.
In particular, the belief space is far too large to either apply naive planning or exact inference.
Instead, we propose to combine: a specialized POMCP algorithm, a particle filtering approximation technique, and a targeted prior construction.

First, we train networks on the priors we have over the unknown dynamic terms (\emph{e.g.}, $\{ w_{hT} \} \approx \humanPrior$). 
Then we initialize our belief $b_0$ by sampling particles $(s, w_{hT}, w_{hO})$ from our joint prior ($s \sim p_{s_0}$, and $(w_{hT}, w_{hO}) \sim  \{ w_{hT}, w_{hO} \}$).
At each time step, our planner builds a look-ahead tree, as POMCP does, but where each simulation starts with sampling an augmented state $\sim b$ and using dynamics in~\cref{eq:main} to generate trajectories.
For belief tracking, we use a new type of rejection sampling.
This algorithm samples particles $(s, w_{hT}, w_{hO})$, proposes the next augmented states (again with dynamics in~\cref{eq:main}) and accepts those that generated the perceived observation.

This approach, first, can do approximate belief tracking through particle filtering in our specialized Bayes model and, second, pick actions efficiently with respect to our most up-to-date understanding of the environment.
The result is a quick and feasible online algorithm that manages to both be sample efficient and capable of exploiting expert knowledge directly into the initial belief in a principled manner.

\section{EXPERIMENTS}

We experiment on two human-robot interaction domains in which the agent needs to learn the tool order and working speeds of human collaborators that cannot be directly observed.
Fortunately (and per usual), the robots come with extensive simulators and need not learn this part of the dynamics.
However, before deployment, the robots have no idea in what order or how quickly these tools must be delivered.
Instead, they have a uniform prior over these unknowns and must learn these factors online. 

\subsection{Ordered-Tools-Delivery (\texttt{OTD})}

A Turtlebot must deliver $T$ tools to a human in an unknown order to complete an assembly task (\cref{fig:domain}a). Turtlebot can carry multiple tools in its basket at once, and the tools are stored in a tool room while the human works in a work room. Each time the human worker receives a correct tool, he needs a fixed number of timesteps to use it before needing another. When a correct tool is delivered, the human's working step (located in the top right corner) increases by one.

\noindent \textbf{State.} 
A state includes the 2D coordinate $x_{coord}$ of Turtlebot, the tools currently carried $t_{carry}$, and the currently needed tool $t_{need}$.
The coordinates are internal and known to the robot, while the next needed tool is not. \emph{Moreover, even $t_{carry}$ is known to the robot, we consider it as a component of $s_h$ because $t_{carry}$ cannot be determined just using the previous $(x_{coord}, t_{carry})$ and the action}. For instance, with \textit{Deliver} actions, determining $t_{carry}$ requires extra information, such as whether the human workers are waiting for a tool. Therefore, $s = (s_r, s_h)$, where
\begin{align}
    s_r=x_{coord} \quad s_h = (t_{carry}, t_{need}).
\end{align}

\noindent \textbf{Observation.}
The agent can observe the current room location $x_{room}$ (work or tool room), the tool it is carrying $t_{carry}$, and the current working step of the human $w_{step}$, which is only observable in the work room. 
The location and tools it is carrying are a function of the robot's known internals (\emph{e.g.}, coordinates), whereas the human working step is external, \emph{i.e.}, $o = (o_r, o_h)$, where
\begin{align}
    o_r = (x_{room}, t_{carry}) \quad o_h=w_{step}.
\end{align}
In this experiment, we consider the observation function known a-priori:
\begin{align}\label{eq:o_r}
    O_r(s) & \coloneqq ( x_{room}, t_{carry}) \\
    O_h(s) & \coloneqq \begin{cases}
          w_{step},  & \text{if } x_{room} = \text{work room} \\
          \varnothing & \text{if } x_{room} = \text{tool room}
          \end{cases}    
\end{align}

\noindent \textbf{Actions/Controllers.}
The action space consists of two types of actions. 
\textit{Get-Tool($i$)} with $i \in \{ 0, 1, \dots, T-1\}$ moves Turtlebot to a \emph{tool reception spot} in the tool room, where it can receive tool $i$ from a Fetch robot that would get the tool from a table. 
\textit{Deliver} moves Turtlebot to \emph{a tool delivery spot} in the work room, where the human can take the tool that he needs \emph{if} it is currently carried by the Turtlebot. 

The navigation and tool pickup transitions are known a-priori $(s_r{=}x_{coord} \sim T_r(s_r, a)$, while the human behavior $(s_h{=}(t_{carry}, t_{need}) \sim T_h(s, a)$ is not.

\noindent \textbf{Reward.} The agent is rewarded $+100$ for delivering a tool. To encourage the agent to achieve the task as quickly as possible, each timestep is given a step reward of $-1$. The cost of \textit{Get-Tool($i$)} tools depends on the time it takes for Fetch to pick and place items and for the Turtlebot to navigate.

\noindent \textbf{Episode Initialization. } An episode starts with Turtlebot (not carrying any tool) at the blue spot; all tools are placed on the tool table; and the human begins with $w_{step} = 0$.


\subsection{Speedy-Tools-Delivery (\texttt{STD})}
In this domain, the Turtlebot must deliver tools to two human workers (Human-0 and Human-1) in separate rooms (\cref{fig:domain}b). Importantly, given the same tool, one of the workers works faster than the other. The table now holds two identical sets of tools, each containing three tools, and each human requires one set to complete the task. The tool order is the same for both workers and assumed to be known, \emph{i.e.}, $\text{Tool }0\rightarrow 1 \rightarrow 2$.

\noindent \textbf{State.} In this domain, the state additionally includes a speed bit $b_{speed}$, which is $1$ if Human-0 works faster than Human-1 and $0$ otherwise. Moreover, $t_{need}$ is now the tools needed by \emph{two} humans, and for the same reason described in \texttt{OTD}, $t_{carry}$ is unknown, making $s = (s_r, s_h)$, where
\begin{align}
    s_r = x_{coord} \quad s_h = (t_{carry}, t_{need}, b_{speed}).
\end{align}

\noindent \textbf{Observation.} Unlike \texttt{OTD}, the agent additionally can observe the unused tools on the table $t_{table}$ when it is in the tool room, and $w_{step}$ is the working steps of \emph{two} humans (only observable in the corresponding work rooms). Adding $t_{table}$ to the observation is necessary because there are two identical sets of tools. Consequently, the complete observation is $o = (o_r, o_h)$, where
\begin{align}
    o_r = (x_{room}, t_{carry}) \quad o_h=(w_{step}, t_{table}).
\end{align}
Like before, we also assume a known observation function:
\begin{align}\label{eq:otd_o_r}
    O_r(s) & \coloneqq ( x_{room}, t_{carry}) \\
    O_h(s) & \coloneqq \begin{cases}
          (w_{step}, \varnothing),  & \text{if } x_{room} = \text{work room} \\
          (\varnothing, t_{table}),  & \text{if } x_{room} = \text{tool room}
          \end{cases}
\end{align}

\noindent \textbf{Action/Controllers.} Similar to \texttt{OTD}, \textit{Get-Tool($i$)} with $i \in \{ 0, 1, 2\}$ will get tool $i$ from the table. However, \textit{Deliver($j$)} with $j \in \{0, 1\}$ will deliver tools for Human-$j$ by moving to the delivery spot of the corresponding work room. Like before, the navigation and tool pick-up transition is known $(s_r{=}x_{coord} \sim T_r(s_r, a)$, whereas the human factors $(s_h{=}(t_{carry}, t_{need}, b_{speed}) \sim T_h(s, a)$ must be learned.

\noindent \textbf{Reward.} We apply the same reward for a correct tool delivery and use the same cost for \textit{Get-Tool($i$)}. Apart from a step reward of $-1$, a step penalty of $-30$ and $-10$ are incurred if the faster and the slower human must wait, respectively. These additional rewards are necessary for the agent to learn to prioritize the faster worker.

\noindent \textbf{Episode Initialization. } Like \texttt{OTD}, Turtlebot starts at the blue spot with no tool. The two tool sets are placed on the table, and the two humans begin with $w_{step} = \{ 0, 0 \}$.

\subsection{Priors}

In \texttt{OTD}, the transition prior assumes a uniform distribution over $T!$ possible delivery orders for $T$ tools. Assuming a \emph{correct} observation model, we represent these priors using $T!$ parameter sets $(w_{hT}, w_{hO})$. To generate training data for training a parameter set, we pair a random tool order with the correct observation model to sample the dynamics. We then use a random policy to interact with the resulting dynamics to generate the data.

In \texttt{STD}, we consider a set $S$ consisting of $|S|$ possible working speeds. Because the speeds are different (one human worker works faster than the other worker),  $|S|^2 - |S|$ possible different speed combinations are possible. Over these combinations, we assume a uniform transition prior, which is defined using $|S|^2 - |S|$ parameter sets $(w_{hT}, w_{hO})$. Similar to \texttt{OTD}, to generate training data for training a parameter set, we pair a random speed combination with a \emph{correct} observation model and the \emph{correct} tool order to create the dynamics and then use a random policy to interact.

\subsection{Real-World Experiments}

\begin{figure}[htbp]
    \centering
    \includegraphics[width=0.9\linewidth]{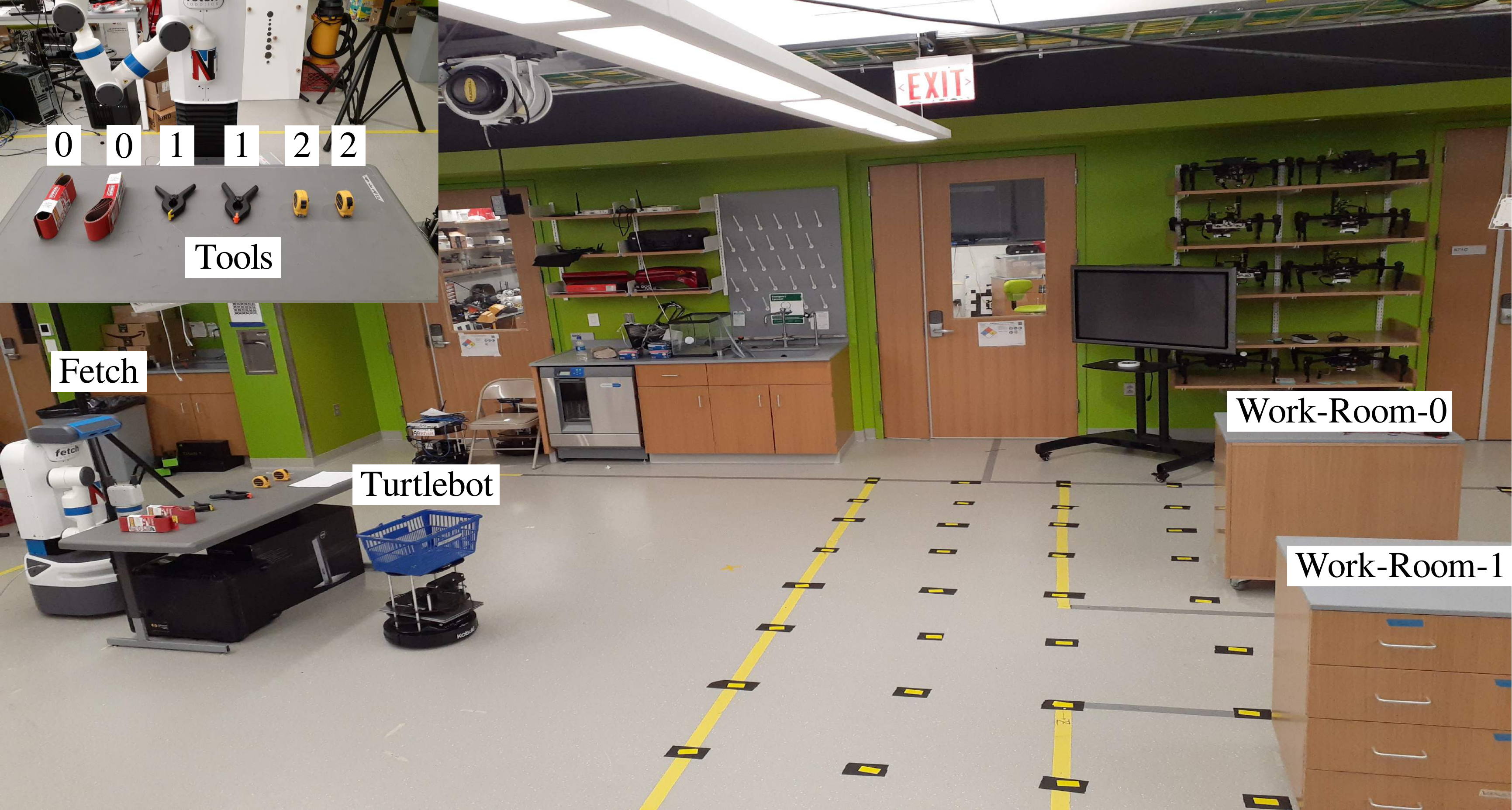}
    \caption{The lab workspace to perform experiments.}
    \label{fig:workspace}
\end{figure}

\noindent \textbf{Set-up.} The experiments take place in a rectangular workspace that measures $5.0 \times 7.0$ meters and contains two tables that represent work rooms (see~\cref{fig:workspace}). Two identical sets of tools are available, each including a clamp, a sandpaper package, and a tape measure. Only one set of tools is used for the \texttt{OTD} task.

\noindent \textbf{Experiment Scenarios.} We perform experiments in two scenarios for each domain:
\begin{itemize}
    \item \texttt{OTD}. The number of tools $T=3$ and the correct tool orders are Tool $0 \rightarrow 1 \rightarrow 2$ and Tool $0 \rightarrow 2 \rightarrow 1$.
    \item \texttt{STD}. The set of possible speeds $S = \{ 10, 20, 30\}$, and the true speeds are $(10, 20)$ and $(20, 10)$.
\end{itemize}

\noindent \textbf{Evaluation Metrics. } We report the mean discounted return with $\gamma = 0.95$, averaged over five runs. Each run lasts 10 and 20 episodes for \texttt{OTD} and \texttt{STD}, respectively. We also visualize the one standard error area around the mean.

\begin{figure}[ht]
  \centering
  \begin{subfigure}[t]{0.47\linewidth}
    \includegraphics[width=\linewidth]{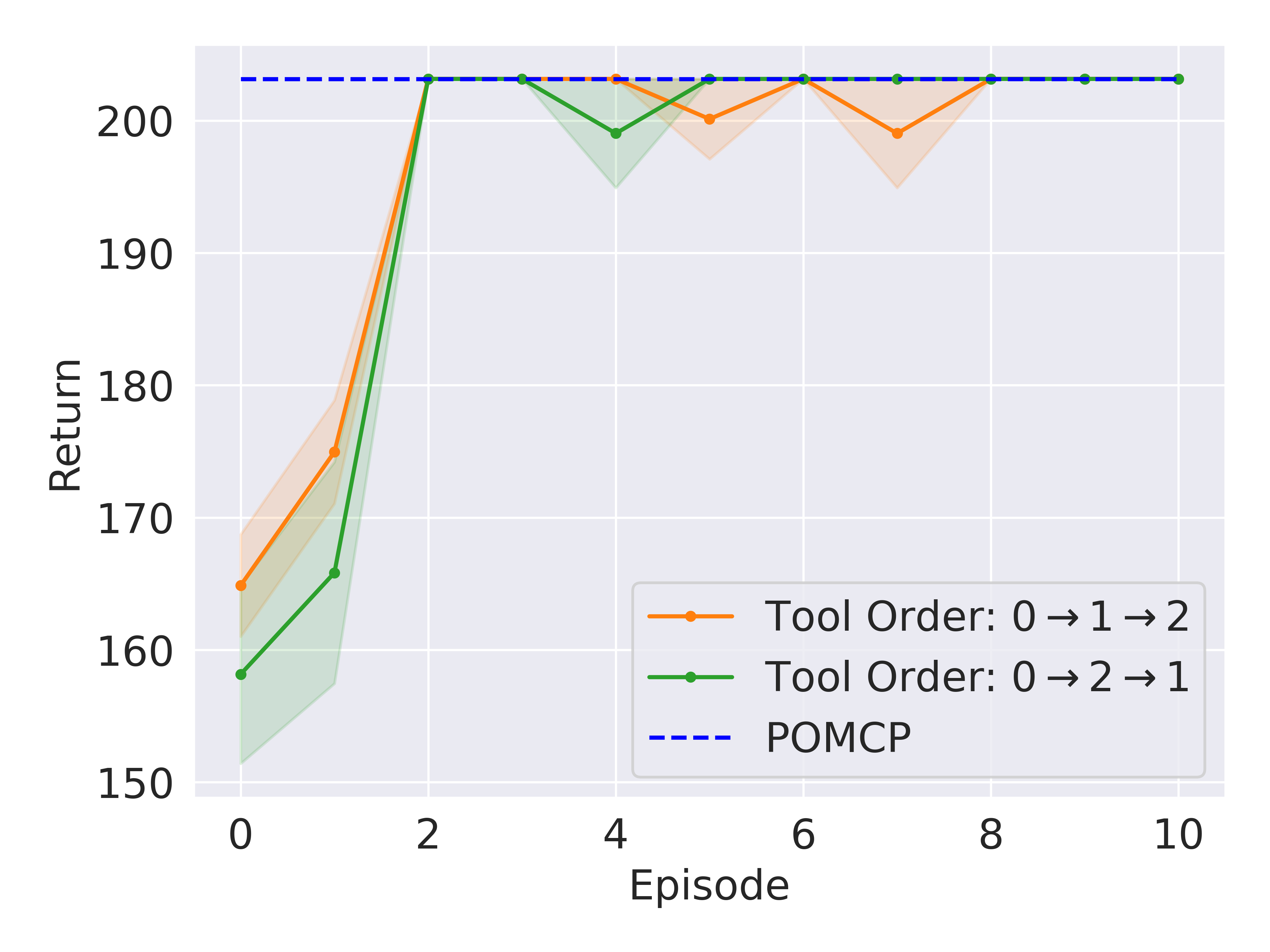}
    \caption{\texttt{OTD} Real} \label{fig:otd-real}
  \end{subfigure}
  \begin{subfigure}[t]{0.47\linewidth}
    \includegraphics[width=\linewidth]{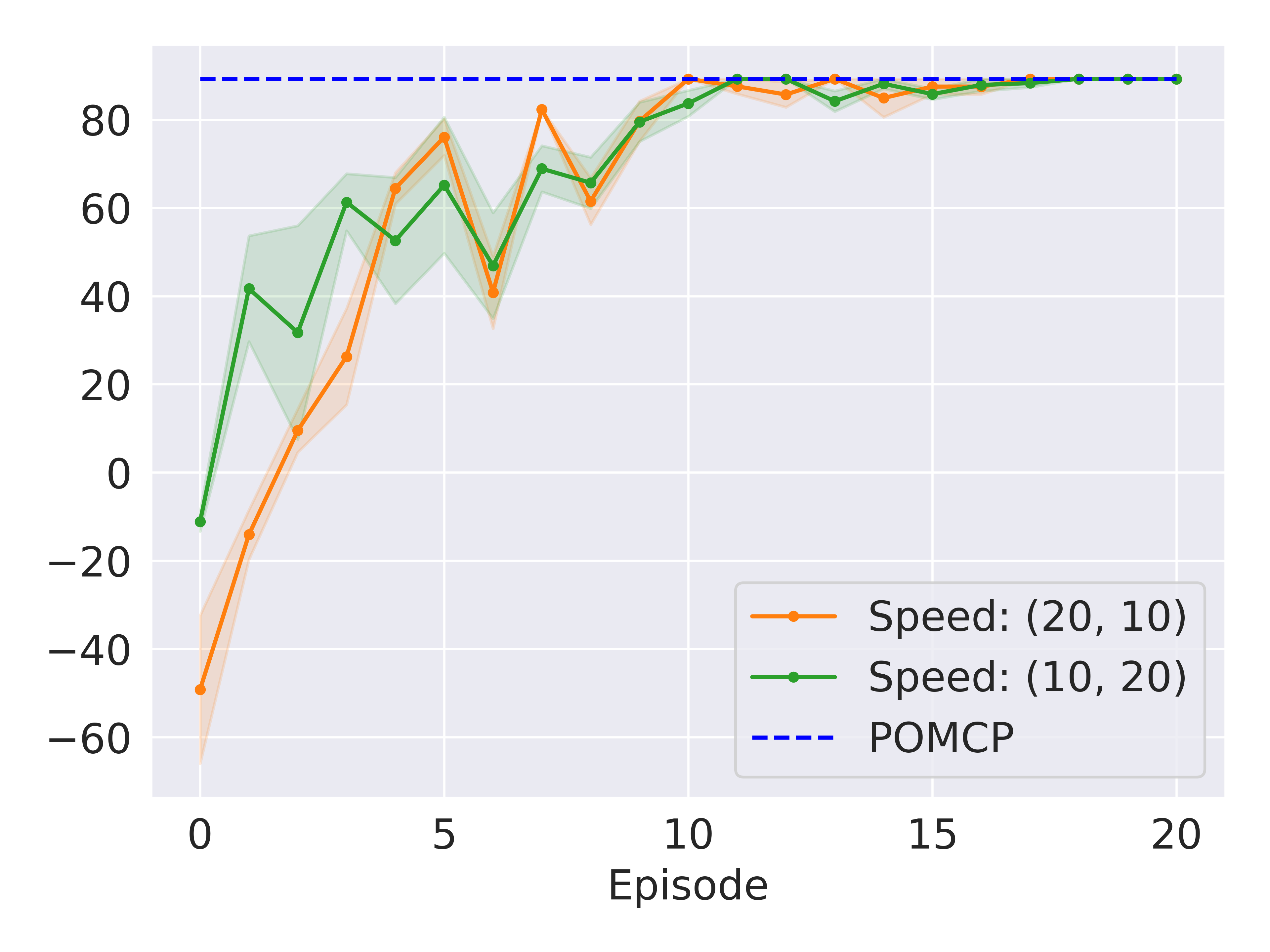}
    \caption{\texttt{STD} Real} \label{fig:std-real}
  \end{subfigure}
\caption{Real-world results in \texttt{OTD} with three tools and \texttt{STD} with two speed combinations. The dotted lines indicate the upper bound of POMCP~\cite{silver2010monte} using the \emph{true} POMDP. Results averaged over five seeds with shaded one standard error.}
\label{fig:results}
\end{figure}

\noindent \textbf{Results.} \Cref{fig:otd-real} shows that our method can reach near-optimal performance within ten episodes in both domains. Such performance will unlikely be achievable by pure RL methods (see a comparison in~\cref{sect:rl_baselines}). We will later show in~\cref{sec:abl} that our approach also outperforms BADDr, given the same amount of initial training for the prior networks.

In \texttt{OTD}, our agent nearly reaches the performance of POMCP running on the \emph{true} POMDP after three episodes. During the first episode of \texttt{OTD}, the agent relies solely on the initial prior, as the dynamics parameters have not yet been updated. The observed behavior of the agent is to gather all three tools and then deliver them. While this strategy yields a reasonable outcome, it is sub-optimal as the worker only requires a tool once they have finished using the current one. In the second episode, the agent performs better by only taking two tools in two consecutive actions, delivering them, and then returning later for the third tool. In the third episode, it learns to retrieve one tool at a time and deliver it until all tools have been supplied, which is optimal. In order to see the robot in action, please see our video.

\texttt{STD} seems more challenging as the agent needs more episodes to perform well. Specifically, \cref{fig:std-real} indicates that our agent can reach the performance of POMCP within ten episodes by delivering all tools in the correct order for both tested working speed combinations. The final policy involves taking two tools of the same type (\emph{e.g.}, tool 0) and delivering them one by one, starting with the faster worker and then moving on to the slower worker. This strategy allows the agent to leverage the close distance between the two work rooms. Please see our video for the learned policy.

\noindent \textbf{Simulating $T_r(s_r, a, s'_r)$. }We first build a map of the experiment area and import it to a commonly used simulator of Turtlebot in Gazebo. This simulator then acts as $T_r$, where we can set the 2D coordinate $x_{coord}$ of Turtlebot, send moving actions, and retrieve the next coordinates $x'_{coord}$.

\noindent \textbf{Obtaining Observations. } Tools on table $t_{table}$ are determined using the RGB image of the tool table taken by Fetch's head camera. The human workers' working steps $w_{step}$ are the number of times the minimum depth in the image from Turtlebot's depth camera falls below a certain threshold, indicating that a human is approaching to pick up a tool. The room location $x_{room}$ is determined by comparing $x_{coord}$ with the rooms' dimensions. During an episode, $t_{carry}$ is calculated by keeping track of $t_{table}$ and $w_{step}$.

\noindent \textbf{Other Implementation Details.} \textit{Get-Tool($i$)} actions of Fetch are pre-recorded using a multi-step process. First, point cloud data from the head camera is projected into an OpenRAVE~\cite{diankov2008openrave} environment. Then, the OMPL~\cite{sucan2012open} library is used for motion planning. Fetch and Turtlebot are controlled through a ROS\cite{quigley2009ros} node, which sends observations to a planning node via a ROS service. The planning node, allowed $N_s = 1024$ simulations, responds with the computed action after about 2.5 seconds of planning. The prior networks implemented in PyTorch are three-layered with 128 neurons per layer and \texttt{Tanh} activation function with a dropout rate of 0.1. They are trained for 2,000 epochs using SGD with a learning rate 0.1, then switched to 0.001 for online learning.

\subsection{Non-Bayesian RL Comparison in Simulation}
\label{sect:rl_baselines}
\noindent \textbf{Baselines.} To further investigate the sample efficiency of our method, we compare it against several non-Bayesian RL baselines in \emph{simulation}: \underline{DRQN}~\cite{hausknecht2015deep} is a recurrent version of Deep Q-Networks (DQN)~\cite{mnih2015human} and is a classical baseline for POMDPs. \underline{R-PPO} is the recurrent version of Proximal Policy Optimization (PPO)~\cite{schulman2017proximal}. Discriminative Particle Filter RL (\underline{DPFRL})~\cite{ma2020discriminative} is a model-based RL POMDP method that performs reinforcement learning on the features from a differentiable particle filter. \underline{DreamerV2}~\cite{hafner2020mastering} is a strong model-based agent which alternates between learning a \emph{recurrent} world model (therefore can tackle POMDPs) and performing RL using imagined data. It outperformed strong model-free baselines such as~\cite{hessel2018rainbow} in the Atari benchmark. Finally, like before, we also include \ul{POMCP} running on the \emph{true} POMDP as an upper bound performance.

\begin{figure}[ht]
  \centering
  \begin{subfigure}[t]{0.47\linewidth}
    \includegraphics[width=\linewidth]{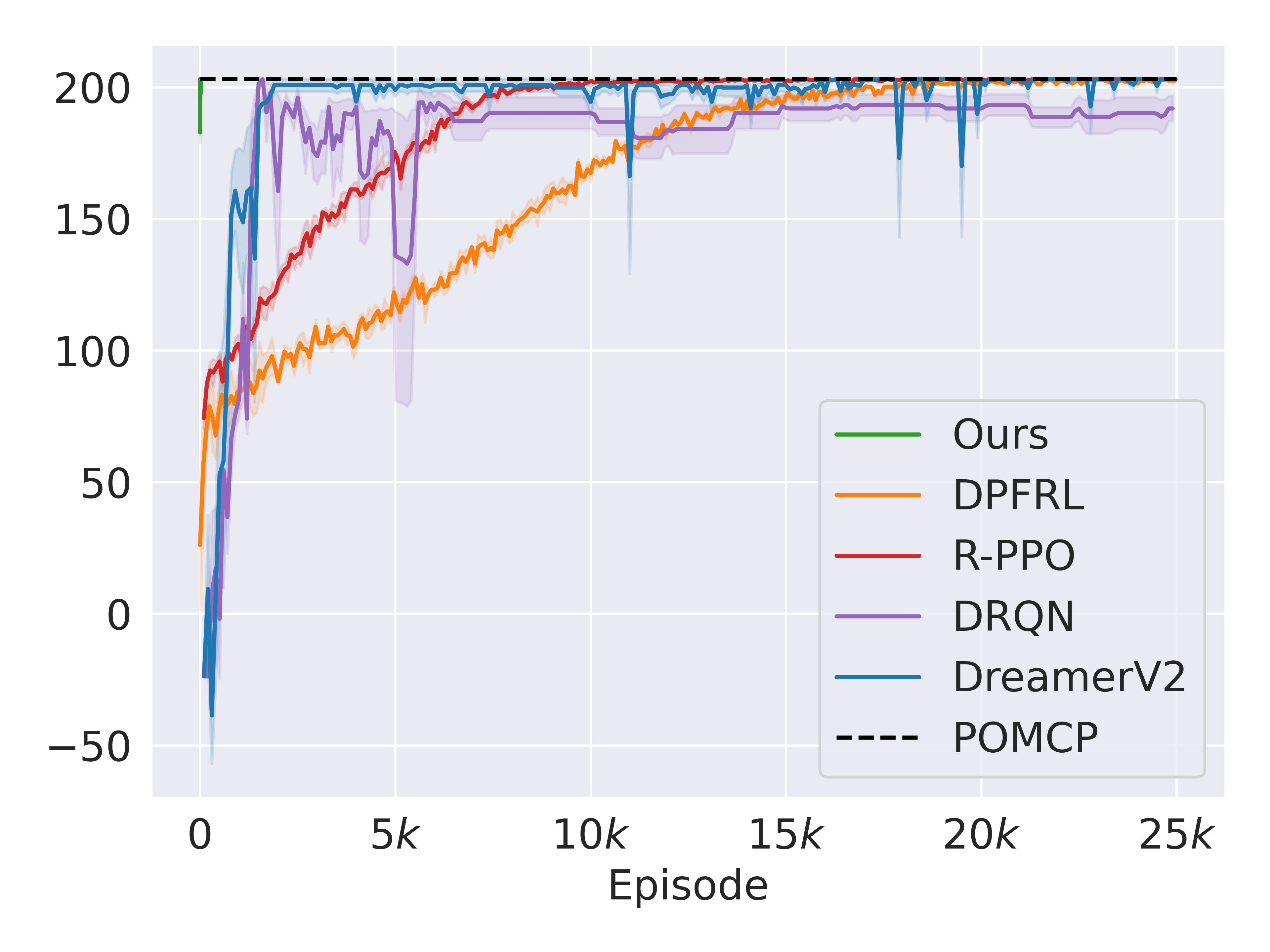}
    \caption{\texttt{OTD} Sim} \label{fig:otd-rl}
  \end{subfigure}
  \begin{subfigure}[t]{0.47\linewidth}
    \includegraphics[width=\linewidth]{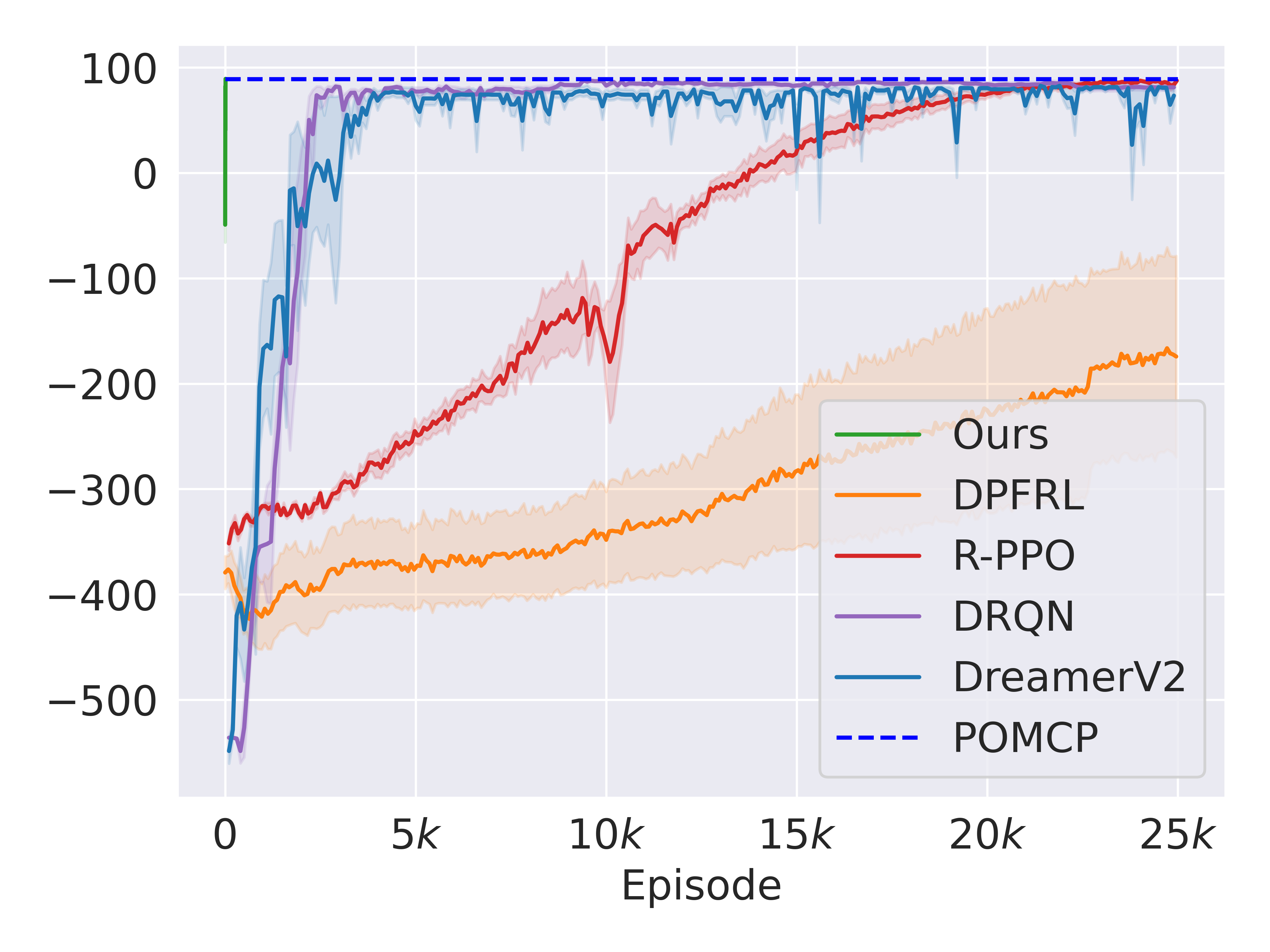}
    \caption{\texttt{STD} Sim} \label{fig:std-rl}
  \end{subfigure}
\caption{Simulation results against RL baselines. Averaged five seeds with shaded one standard error.}
\label{fig:rl_results}
\end{figure}

\noindent \textbf{Results.} In~\cref{fig:rl_results}, our method (barely seen in the top left corner) exhibits significantly greater sample efficiency compared to RL baselines in both domains. Additionally, our approach outperforms from the beginning due to leveraging prior information. Specifically, in the \texttt{OTD} domain, DreamerV2 requires approximately 1,000 episodes to achieve the performance our method attains in just five episodes. In the more challenging \texttt{STD} domain, the numbers are 2,500 and 10, respectively. Among baselines, DreamerV2 performs best in \texttt{OTD} while DRQN performs best in \texttt{STD}. Regardless, these baselines sometimes act suboptimally at the end of the training (see spikes in their learning curves). For instance, in \texttt{OTD}, DreamerV2 still sometimes outputs a redundant \textit{Get-Tool} action instead of delivering its current tool to the waiting human. And, in \texttt{STD}, DRQN occasionally prioritizes the slower human worker first. In contrast, R-PPO exhibits better convergence performance in both domains. 

\subsection{Ablation Studies in Simulation}
\label{sec:abl}

To ablate our agent, we perform experiments in simulation with the \texttt{OTD} task with three tools.

\begin{figure}[ht]
  \centering
  \begin{subfigure}[t]{0.47\linewidth}
    \includegraphics[width=\linewidth]{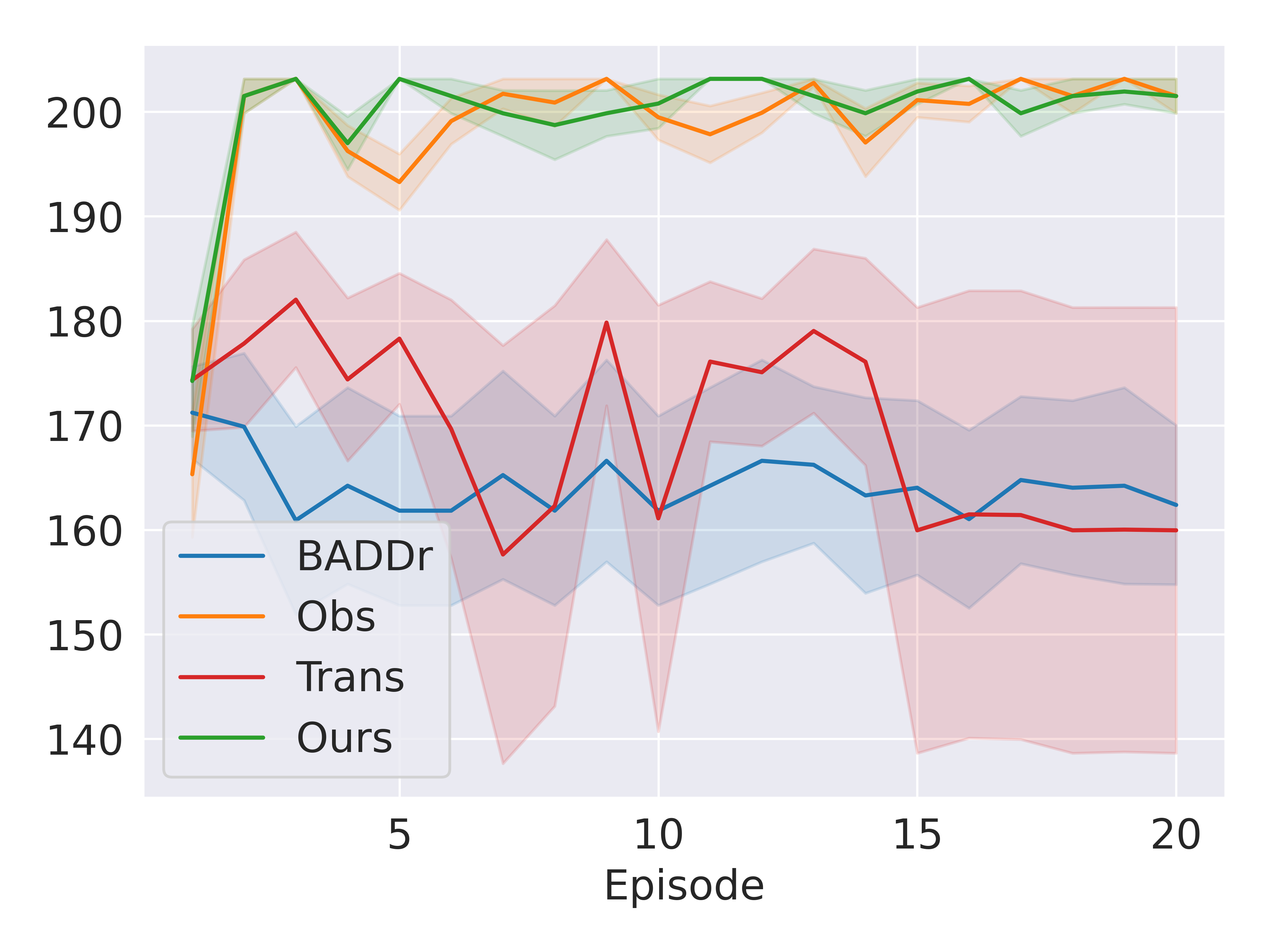}
    \caption{Effect of Factored Models} \label{fig:abl_reduction}
  \end{subfigure}
  \begin{subfigure}[t]{0.47\linewidth}
    \includegraphics[width=\linewidth]{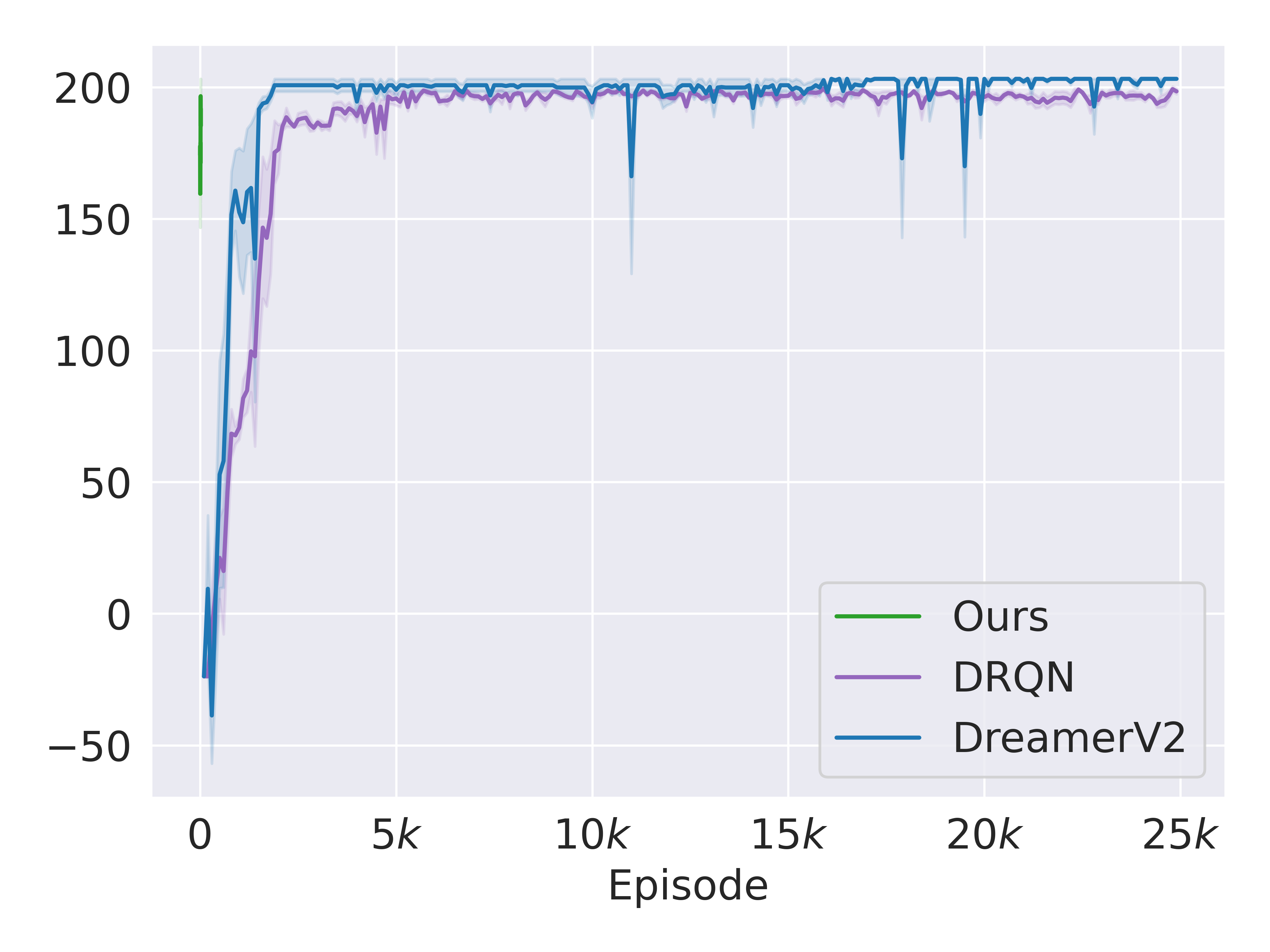}
    \caption{Imperfect Obs. Model} \label{fig:abl_imperfect_obs}
  \end{subfigure}
  \caption{Ablation studies in \texttt{OTD} with three tools.}
\end{figure}

\noindent \textbf{Effect of Factored Models.}
We conducted an experiment where we provided the same initial training for the prior networks (all trained for 2,000 epochs) and compared the proposed approach (\ul{Ours}) with its variants, namely, no factorizations used (\emph{i.e.}, the original \ul{BADDr}), factored transition model only (\ul{Trans}), and factored observation model only (\ul{Obs}). \Cref{fig:abl_reduction} indicates that our method and Obs perform best. Obs likely performs well because $O_r$ in \texttt{OTD} is relatively small, resulting in a slight performance improvement when using the factored observation model. Conversely, we anticipate a more substantial performance improvement in domains where $O_r$ is a significant component.

\noindent \textbf{Using Imperfect Observation Models.} Here, we investigate how our approach performs against DRQN and DreamerV2 (best RL baselines in~\cref{sect:rl_baselines}) when we do not assume a correct observation model. For this purpose, we add \emph{stochasticity} while observing the currently carried tools $t_{carry}$, which is now only correctly observable with a probability $p_{correct}=0.85$. However, the data for initially training the prior observation model is obtained with $p_{correct}=0.5$. In this case,~\cref{fig:abl_imperfect_obs} still confirms the sample efficiency superiority of our approach over the baselines.

\section{CONCLUSIONS}

To our knowledge, this work presents the first on-robot Bayesian RL method for partially observable scenarios. By employing factored dynamics and mixed observability assumptions, our method can rapidly acquire high-quality behavior in long-horizon, real-world tasks within a minimal number of episodes, outperforming pure RL approaches. Although our tasks are relatively simple, the results clearly demonstrate the potential of a Bayesian approach for effective learning directly on physical hardware. One limitation of the approach is a slow inference speed, as the agent needs time to search for the next action. This weakness can be overcome using an RL agent to mimic our agent's actions, as previously investigated in~\cite{guo2014deep}. 

\section*{ACKNOWLEDGMENTS}
We thank Trung-Hieu Nguyen, Minh Nguyen, Van Anh Tran, and Yunus Terzioglu for helping in the robot experiments.
This work is supported by the Army Research Office under award number W911NF20-1-0265 and
NSF grant 2024790.

\bibliographystyle{IEEEtran}
\bibliography{refs}

\end{document}